\documentclass{article}

\usepackage{PRIMEarxiv}

\usepackage[utf8]{inputenc} 
\usepackage[T1]{fontenc}    
\usepackage{hyperref}       
\usepackage{url}            
\usepackage{booktabs}       
\usepackage{amsfonts}       
\usepackage{nicefrac}       
\usepackage{microtype}      
\usepackage{lipsum}
\usepackage{graphicx}
\graphicspath{{media/}}     

\usepackage{multirow}
\usepackage{multicol}
\usepackage{rotating}
\usepackage{pifont}
\usepackage{booktabs}
\usepackage{amsmath}
\usepackage{graphicx}
\usepackage{natbib}
\usepackage{float}

\newcommand{\cmark}{\ding{51}}

\title{From Lengthy to Lucid: A Systematic Literature Review on NLP Techniques for Taming Long Sentences}

\author{
 Tatiana Passali \thanks{Corresponding author} \\
  Aristotle University of Thessaloniki \\
  Thessaloniki, Greece \\
  \texttt{scpassali@csd.auth.gr} \\
 \And
 Efstathios Chatzikyriakidis \\
  Medoid AI \\
  Thessaloniki, Greece \\
  \texttt{stathis.chatzikyriakidis@medoid.ai} \\
 \And
 Stelios Andreadis \\
  Pfizer Center for Digital Innovation \\
  Thessaloniki, Greece \\
  \texttt{Stelios.Andreadis@pfizer.com} \\
 \And
 Thanos G. Stavropoulos \\
  Pfizer Center for Digital Innovation \\
  Thessaloniki, Greece \\
  \texttt{Thanos.Stavropoulos@pfizer.com} \\
 \And
 Anastasia Matonaki \\
  Pfizer Center for Digital Innovation \\
  Thessaloniki, Greece \\
  \texttt{Anastasia.Matonaki@pfizer.com} \\
 \And
 Anestis Fachantidis \\
  Medoid AI \\
  Thessaloniki, Greece \\
  \texttt{anestis@medoid.ai} \\
 \And
 Grigorios Tsoumakas \\
  Aristotle University of Thessaloniki \\
  Thessaloniki, Greece \\
  \texttt{greg@csd.auth.gr} \\
}

\begin{document}
\maketitle
\begin{abstract}
Long sentences have been a persistent issue in written communication for many years since they make it challenging for readers to grasp the main points or follow the initial intention of the writer. This survey, conducted using the PRISMA guidelines, systematically reviews two main strategies for addressing the issue of long sentences: a) sentence compression and b) sentence splitting. An increased trend of interest in this area has been observed since 2005, with significant growth after 2017. Current research is dominated by supervised approaches for both sentence compression and splitting. Yet, there is a considerable gap in weakly and self-supervised techniques, suggesting an opportunity for further research, especially in domains with limited data. We also observe that despite their potential, Large Language Models (LLMs) have not yet been widely explored in this area. In this survey, we categorize and group the most representative methods into a comprehensive taxonomy. We also conduct a comparative evaluation analysis of these methods on common sentence compression and splitting datasets. Finally, we discuss the challenges and limitations of current methods, providing valuable insights for future research directions. This survey is meant to serve as a comprehensive resource for addressing the complexities of long sentences. We aim to enable researchers to make further advancements in the field until long sentences are no longer a barrier to effective communication. 
\end{abstract}

\section{Introduction}
\label{sec:introduction}
Several studies going back more than 100 years, have shown that the length of a sentence in words is negatively correlated with its readability~\cite{Dubay2006TheStudies}. The Oxford guide to plain English recommends sentences of 15 to 20 words~\cite{Cutts2007OxfordEnglish}. According to the cognitive load theory~\cite{sweller2011cognitive}, humans' limited working memory capacity makes it difficult to process longer and more complex sentences. Such sentences make it difficult for readers to comprehend the meaning and intention of the writer. 

However, people tend to include long and complex sentences in their writing. Therefore, there is an increasing need for automated systems that can effectively deconstruct long sentences into shorter and more comprehensive ones. This is critical in many societal applications such as social media, marketing as well as accessibility services, where long sentences make the content much harder to digest. In addition, shorter sentences can benefit people with reading disabilities and improve many natural language processing (NLP) applications, such as text summarization~\cite{lin2003improving, zajic2008, li2013document}, headline~\cite{Rush2015ASummarization, Filippova2015} and subtitle generation~\cite{luotolahti2015sentence}, machine translation~\cite{li2020explicit} and speech processing~\cite{wu2007,wang2012two}.

Transforming a long sentence into a more concise form without losing any important information is not always a straightforward task. There are two ways for addressing this problem: a) shortening the long sentence, without losing key information from it, and b) splitting it into two or more short sentences. The first task, which is met in the literature as {\em sentence compression} or {\em sentence summarization} receives a long sentence as input and generates a shortened version retaining the most important information. Sentence compression can be extractive or abstractive. Extractive methods work by simply deleting words from the input sentence in contrast to abstractive ones which involve more complex paraphrasing, rewriting, and reordering operations. Studies of human summarizers have shown that during compression, besides deleting words, they commonly also do paraphrasing, generalization, and reordering~\cite{Jing2000}. On the other hand, human-written reference summaries have been found to exhibit high word overlap with the source sentence, even preserving word order to a large extent~\cite{Schumann2020DiscreteExtraction}. 
Sentence splitting segments a given long sentence into two or more shorter sentences of equivalent meaning. We can also meet sentence splitting in the literature as {\em split and rephrase} task~\cite{narayan-etal-2017-split, aharoni-goldberg-2018-split, Zhang2020, kim-etal-2021-bisect}. The split and rephrase task does not involve any deletion or simplification, apart from the required syntax modification of the shorter sentences. In many prior studies, sentence splitting is typically met as a preprocessing step within machine translation~\cite{lee2008transformation}, discourse relation~\cite{yu2019gumdrop} or sentence simplification systems~\cite{siddharthan2014survey}.

{\em Sentence simplification} is a related task whose goal is to modify the content and structure of a sentence in order to make it {\em easier to understand}, while retaining its main idea and most of its original meaning. Sentence simplification does not directly target the reduction of sentence length; instead, its primary focus lies in enhancing comprehension by simplifying the content of the sentence. In many cases, length reduction might naturally occur as a consequence of content simplification either achieved with word deletion or simplifying paraphrasing operations. This survey focuses on the simplification of a sentence explicitly through length reduction by employing sentence compression or splitting methods. 

\begin{figure}
    \centering
    \includegraphics[width=0.6\textwidth]{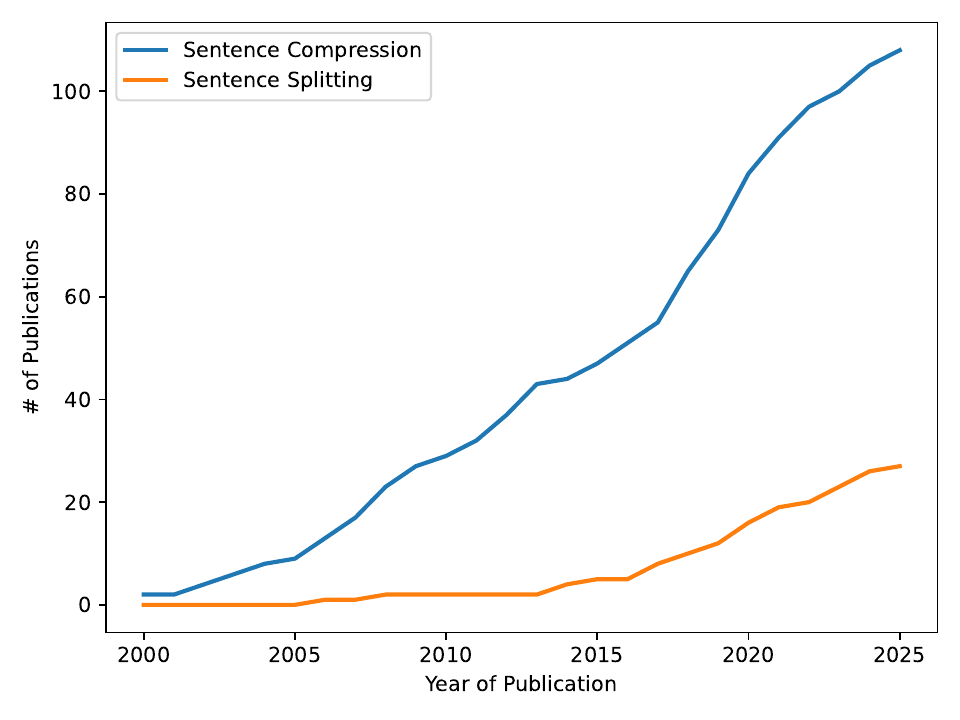}
    \vspace{1.5em} 
    \caption{Cumulative number of publications from 2000 to 2025 for sentence compression and sentence splitting, demonstrating the momentum and growing interest in the field.}
    \label{fig:trend_of_interest_without_landmarks}
\end{figure}

Addressing long sentences has been a subject of research in the literature for over 20 years, as shown in Fig.~\ref{fig:trend_of_interest_without_landmarks}, where the number of publications related to sentence compression and splitting from 2000 to 2025 are presented, as discovered through our systematic survey methodology (see Section~\ref{sec:methodology}).
We can see that there has been a growing interest in this subject since 2005, which has intensified after 2017. Despite the increasing research interest, there are limited surveys in the field. Those are mainly focused on extractive sentence compression~\cite{fathima2018review} and evaluation of sentence compression methods~\cite{napoles2011evaluating}. This paper provides a comprehensive review that explores both sentence compression and sentence splitting for resolving long sentences. To the best of our knowledge, this is the first systematic review on this topic.

The article contributions can be summarized as follows:

\begin{itemize}
    \item We provide a systematic review of state-of-the-art methods for dealing with long sentences, including both sentence compression and sentence splitting.
    \item We categorize the main approaches with the most representative methods into a comprehensive taxonomy.
    \item We provide a comparative analysis between existing methods on widely used datasets for this task.
   \item We discuss the challenges and limitations of existing methods.
\end{itemize}

The rest of this paper is organized as follows. Section~\ref{sec:methodology} presents an overview of the survey methodology that we followed based on the PRISMA guidelines. Section~\ref{sec:sentence_compression} and Section~\ref{sec:sentence_splitting} discuss sentence compression and sentence splitting, respectively, including methods, datasets and a comparative evaluation. Section~\ref{sec:discussion} discusses the challenges and limitations of sentence compression and sentence splitting while Section~\ref{sec:conclusions} concludes our work. 

\section{Survey Methodology}
\label{sec:methodology}
This survey was conducted following the PRISMA guidelines~\cite{page2021prisma} for reporting systematic reviews. In this section, we provide an overview of the survey methodology, including eligibility criteria, information sources, and search strategy, as well as the PRISMA flowgram and the results of the study selection.

All original works concerning the task of dealing with long sentences were considered eligible for this systematic review. The inclusion criteria adopted were: i) publication date between 2000 and 2025; ii) written in the English language; iii) works providing a full-text article; and iv) journal and conference publications. Unpublished works such as dissertation studies, theses and technical reports were excluded from this review. Due to the large number of works, studies with few citations (less than 10 citations in total) that were published prior to 2017 were also excluded from this review. 

All articles were collected by searching relevant papers in the following databases: i) Google Scholar; ii) Scopus; iii) ACM Digital Library; iv) Springer Digital Library; v) IEEE Explore Digital Library; and vi) ArXiv. In addition, the reference lists of the extracted papers were manually searched for additional relevant papers.

The following search query was used to identify relevant papers: (``sentence summarization'' OR ``sentence compression'' OR ``sentence splitting'' OR ``split and rephrase''). We arrived at this query by starting from (`sentence compression'' OR ``sentence splitting'') and gradually expanding it with additional synonymous keywords that we discovered on the way. We initially searched using all the specified criteria for the years 2000 to 2022, limiting our search to the first 300 records in each database to maintain a manageable result pool. We then repeated this process for the years 2023 to 2025, limiting our search to 30 records per database, as no relevant results were found beyond this threshold.

\begin{figure}
    \centering
    \includegraphics[width=0.85\linewidth]{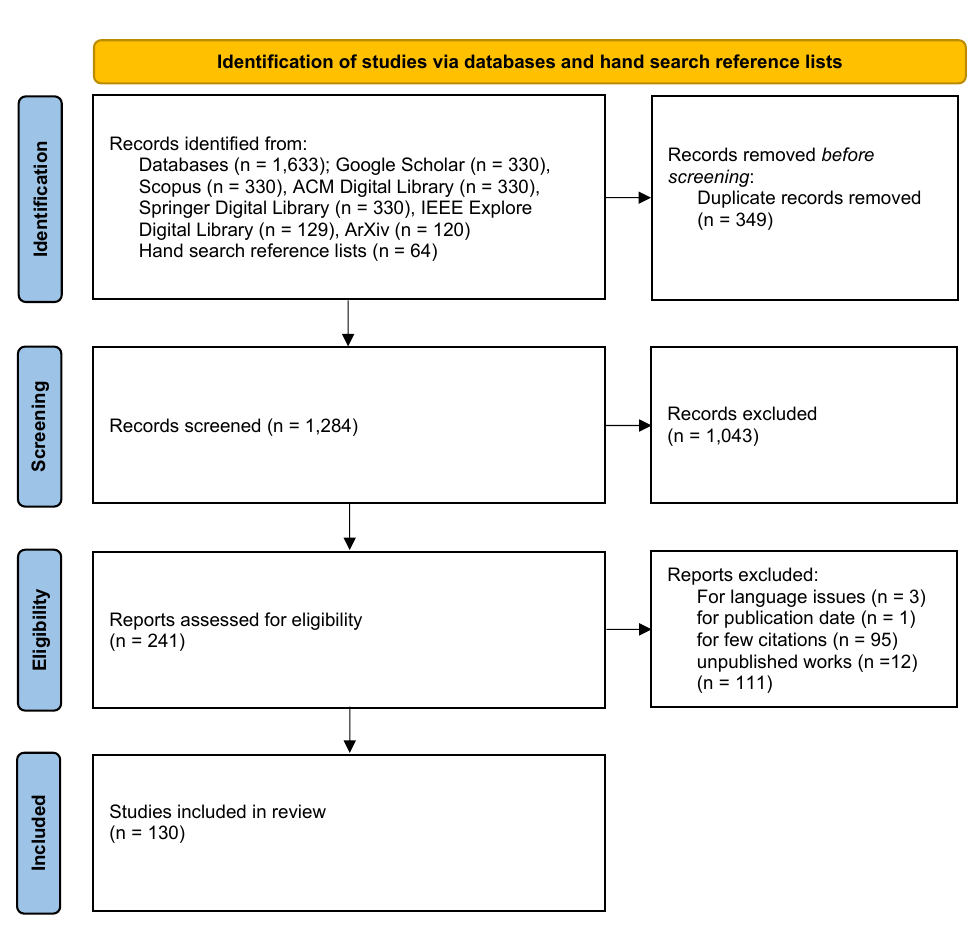}
    \vspace{1.5em}
    \caption{PRISMA flow diagram of the study selection process.}
    \label{fig:PRISMA_flow_diagram}
\end{figure}

A total of 1,633 records were collected via digital databases (Google Scholar n = 330, Scopus n = 330, ACM Digital Library n = 330, Springer Digital Library n = 330, IEEE Explore Digital Library n = 129, ArXiv n = 120) and hand search reference lists (n = 64). 349 records were removed due to duplication before the screening stage. Further 1,043 records were excluded during the title and abstract screening since they did not fall within the scope of this review. After reviewing the full-text, records were excluded for eligibility due to language issues (n = 3), publication date (n = 1), few citations (n = 95) and unpublished works (n = 12). During the study selection, 130 records fully met the initial criteria and are included in this review. The study selection flowgram is shown in detail in Fig.~\ref{fig:PRISMA_flow_diagram}.

\section{Sentence Compression}
\label{sec:sentence_compression}
In this section, we review methods, datasets, and evaluation metrics for the task of sentence compression. In addition, we provide a comparative evaluation of the most representative methods for this task. Finally, we discuss other trends observed in the literature including length control, the increasingly relevant domain of multilingual and cross-lingual sentence compression and the growing interest in multimodal sentence summarization.

\begin{figure}
    \centering
    \includegraphics[width=0.99\textwidth]{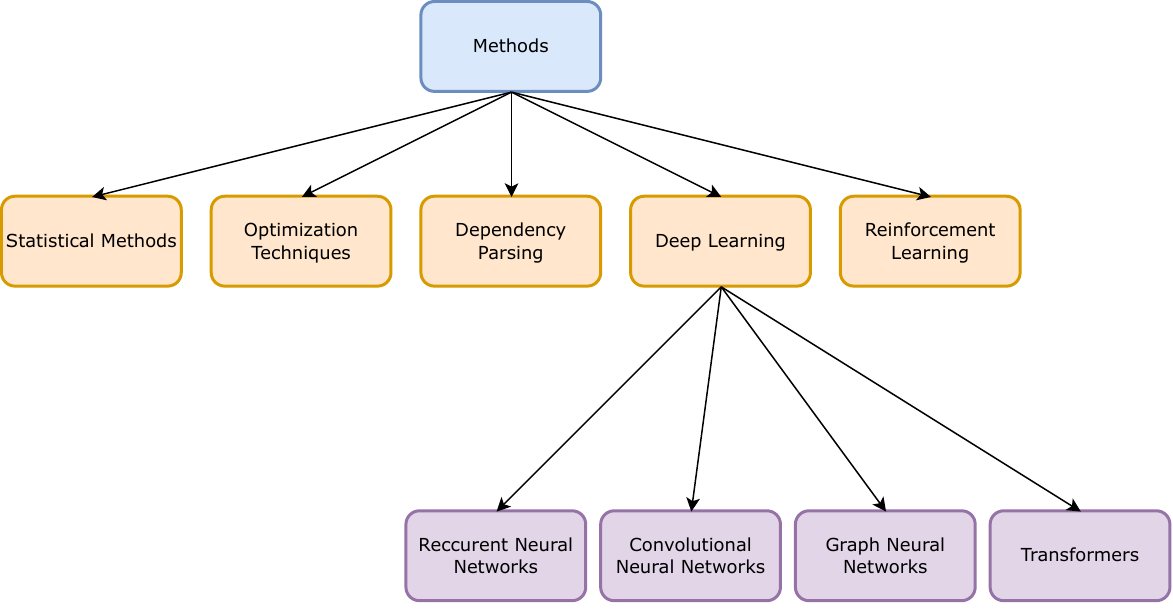}
    \vspace{1.5em}
    \caption{Taxonomy of sentence compression methods}
    \label{fig:compression_taxonomy}
\end{figure}

\subsection{Methods}
\begin{figure}
    \centering
    \includegraphics[width=0.7\textwidth]{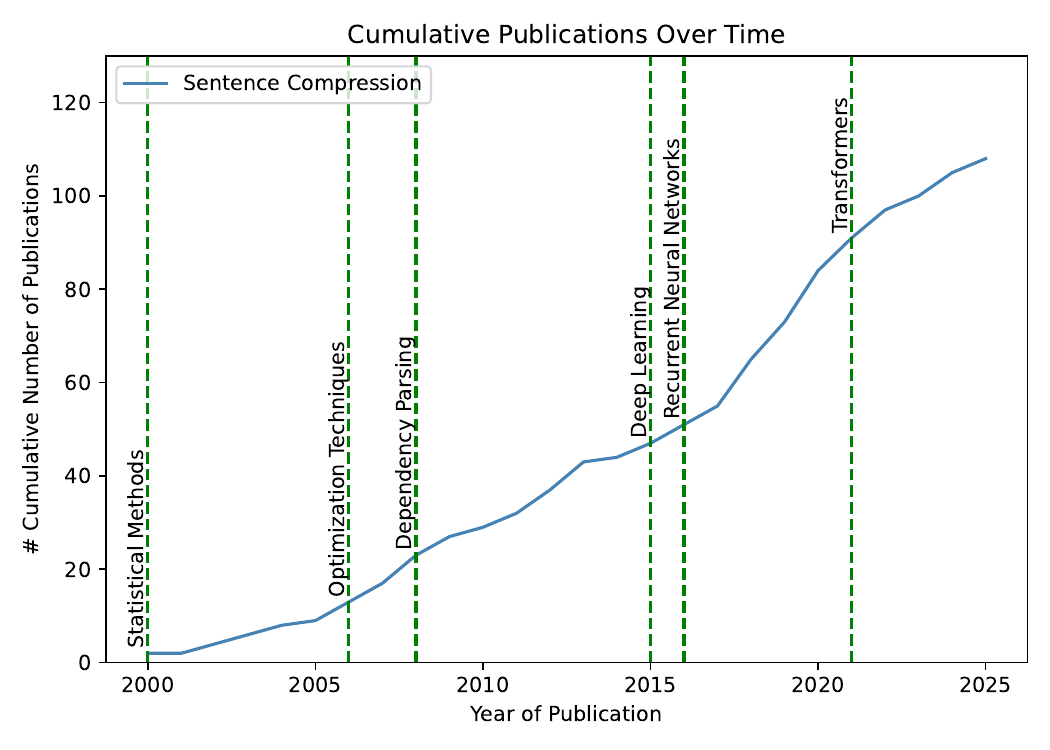}
    \vspace{1.5em}
    \caption{Cumulative number of publications per year from 2000 to 2025 for sentence compression along with landmarks of key technology adoption.}
    \label{fig:sentence_compression_cumulative}
\end{figure}

Sentence compression can be either extractive or abstractive. Extractive methods reduce the length of the input sentence by simply removing words while abstractive methods achieve compression using additional operations besides deleting such as paraphrasing, rewriting and reordering. However, both approaches 
share the same underlying techniques (e.g., statistical methods, reinforcement learning, deep learning). Thus, this review categorizes the methods based on the employed technique. 
Figure~\ref{fig:sentence_compression_cumulative} shows the cumulative number of publications from 2000 to 2024, along with landmarks of key technology adoption. Early works from 2000 to 2005 were limited and primarily relied on statistical methods. However, a significant increase in interest became evident after 2008, especially from 2015 onwards with the introduction of deep learning techniques. A spike in interest is also observed after the emergence of Transformers. The increasing attention in sentence compression methods can be attributed to the remarkable advancements in the field of NLP, as well as the ever increasing textual information overload from various sources. A high-level taxonomy of sentence compression methods is shown in Figure~\ref{fig:compression_taxonomy}.

\subsubsection{Statistical Methods}
Early works on extractive sentence compression involve the use of statistical methods like noisy channels and maximum entropy models. Such methods aim to estimate the joint probability distribution $P(x, y)$ of a given long sentence $x$ and its target compression $y$. Noisy channel models are used to determine which words must be removed from the input sentence~\cite{Knight2000, Knight2002, turner2005supervised, Galley2007}. These models work by adding words to the original short sentence resulting to a noisy long sentence. The goal is to find the compression that maximizes the probability of the short sentence to be the condensed version of the long sentence.~\cite{Knight2000} are the first to employ a noisy channel with a small dataset of 1,067 compressions.~\cite{turner2005supervised} adapt this model for unsupervised learning without the need of parallel data, while other approaches involve additional lexical variables into the model~\cite{Galley2007} or extend it with a maximum entropy model~\cite{unno2006trimming}.

\subsubsection{Optimization Techniques}
Several techniques approach the task of sentence compression as an optimization problem to find the optimal solution or set of solutions from a given set of possible solutions, subject to defined constraints. These approaches include Integer Linear Programming (ILP)~\cite{clarkelapata2006constraint,clarke2007modelling,Clarke2008GlobalApproach, clarkelapata2010discourse, thadani2013-sentence} , Dynamic Programming (DP)~\cite{napoles2011paraphrastic, Liu2022Controlling} or other optimization algorithms such as first-choice hill climbing~\cite{Schumann2020DiscreteExtraction, liu2022learning}. 

Early methods formulate the task of sentence compression as a constrained optimization problem with ILP. ILP is an optimization technique where a linear objective function and decision variables are subject to linear constraints that are represented as linear inequalities. Similar to other optimization problems, ILP aims to find a global optimum solution by searching a given space of possible solutions. The decision variables must be integer values while the objective function is typically formulated as a weighted sum of the decision variables. For example,~\cite{clarkelapata2006constraint} keep the words of the input sentence that maximize a language model's objective function. The scoring function is subject to several hand-crafted grammar and syntactic constraints. \cite{clarke2007modelling} exploit the discourse relations between sentences in the whole input document instead of handling each sentence separately.~\cite{Clarke2008GlobalApproach} set global constraints to find the optimal compression. Unlike the local constraints of the previous approaches, global constraints allow for considering the whole sentence and its relations rather than processing isolated segments and units. This approach includes both supervised, semi-supervised and unsupervised models.~\cite{thadani2013-sentence} use commodity flow constraints that allow for changing the word ordering to ensure the appropriate syntax of the compression, unlike the hard global constraints of~\cite{Clarke2008GlobalApproach} that are restricted to the same token ordering as the input sentence.~\cite{thadani2014approximation} improves the latency of previous methods using an approximate inference algorithm with Lagrangian relaxation.

Other approaches employ DP to break down the task of sentence compression into smaller sub-problems. The final solution is being constructed by combining the solutions to these sub-problems. More specifically,~\cite{napoles2011paraphrastic} introduce a method where paraphrasing and dynamic programming are combined for character-based compression. This method works by replacing existing words in the input with their paraphrases to extract the optimal solution of non-overlapping paraphrases that minimizes the character length of the sentence. Similarly, in~\cite{Liu2022Controlling} the length reduction is imposed as a character-level length constraint, with possible lengths to be divided into buckets.

Recent methods employ a first-choice hill climbing algorithm that starts with an initial candidate solution and iteratively explores local solutions to find a better one. More specifically,~\cite{Schumann2020DiscreteExtraction} introduce an objective function that is based on hill climbing for removing words from the initial sentence. This function is used to generate extractive compressions in an unsupervised way. In~\cite{liu2022learning} the same idea is adopted for generating pseudo-compressions, while a Transformer model is trained on these compressions. 

\subsubsection{Dependency Parsing}
Dependency parsing is the task of extracting syntactic and grammatical relations between each word in a sentence to better analyze its structure. Dependency parsing is widely used to solve the task of sentence compression in both extractive~\cite{Filippova2008, Filippova2013OvercomingCompression, Galanis2010} and abstractive settings~\cite{Cohn2008, Cohn2009, Galanis2011ACompressor}.

Extractive approaches employ a dependency tree to prune unimportant words of the input sentence~\cite{Filippova2008, Filippova2013OvercomingCompression, Galanis2010}. For example,~\cite{Filippova2008} introduce an unsupervised method for sentence compression based on a dependency tree by pruning sub-trees of the original sentence that do not contain important details.~\cite{Filippova2013OvercomingCompression} propose a supervised method based on dependency parsing that builds upon previous unsupervised work~\cite{Filippova2008}. Instead of using pre-computed edge weights, this method employs a linear function over different feature representations (syntactic, structural, semantic, and lexical) to enhance the readability and informativeness of the compression.~\cite{Galanis2010} employ a dependency tree to generate different compression candidates by pruning unimportant words from the source sentence, while a support vector regression (SVR) model is used to select the best candidate. This method can successfully perform extractive compression without the need for defined rules, as in~\cite{Clarke2008GlobalApproach}. 

~\cite{Cohn2008} propose the first abstractive sentence compression model that employs a tree-to-tree rewriting task with additional rewriting operations such as substitution, insertion and reordering, instead of simple deletions.~\cite{Cohn2009} approach this task using a syntactic dependency tree with a large set of features. ~\cite{Galanis2011ACompressor} introduce an abstractive approach with a generate-and-rank sentence model. First, different extractive candidates are generated using a dependency tree, which are then transformed into abstractive candidates with paraphrasing tools. The best candidate is selected based on an SVR ranking module. Similarly,~\cite{cohn2013} build upon rewriting and paraphrasing operations for abstractive sentence compression rather than simply deleting words from the input. 

Even though these approaches can be successful in generating grammatically correct sentences, they might result in pruning important words from the input sentence. Recent approaches~\cite{kamigaito2018higher, kamigaito2020syntactically} claim that a long distance between the root node and the child nodes in a dependency tree should not be neglected. For example,~\cite{kamigaito2018higher} use an LSTM paired with a dependency parser to keep words that belong to the parent dependency chain of the word that should be retained. ~\cite{kamigaito2020syntactically} extend this approach to consider the dependency chain of the children as well. Another approach that employs dependency parsing in combination with neural architectures is introduced in~\cite{park2021sentence}. More specifically, in this work, a neural model for extractive sentence compression is proposed, where the dependency tree of each word of the input sentence is encoded using a graph convolutional network (GCN). Then, a pre-trained BERT model is combined with multiple GCN layers, while a scoring layer decides which words of the input sentence should be retained or removed. 

\subsubsection{Deep Learning}
With the advent of deep learning (DL), several neural extractive and abstractive methods have been introduced. Since sentence compression can be easily formulated as a sequence-to-sequence (seq2seq) tasks, several neural models such as Recurrent Neural Networks (RNNs)~\cite{Yu2018operation, Chopra2016, Liu2017Attention, zhou2017selective, guan2021knowledge}, Convolutional Neural Networks (CNNs)~\cite{hou2020cnn}, Graph Neural Networks (GNNs) and Transformers~\cite{duan2019contrastive, west2019bottlesum, Guan2021Frame} have been proposed. The first neural abstractive approach was introduced in~\cite{Rush2015ASummarization}, where an attention-based contextual encoder combined with a feed-forward neural language model is proposed. However, a feed-forward model might not be capable of capturing all the long-term dependencies of sequential information in an input sentence compared to RNNs.

{\bf RNNs} are among the most common and widely used neural architectures for sentence compression. More specifically,~\cite{Chopra2016} extend the previous feed-forward neural model~\cite{Rush2015ASummarization} with attention-based RNNs, where the decoder is fed with an additional attention-based input during the text generation. Similarly,~\cite{Yu2018operation} propose an abstractive operation network inspired by the well known pointer-generator network~\cite{See2017GetNetworks}. This technique models sentence compression as an editing task where redundant words are removed from the input sentence and new words are created either by rephrasing or copying existing ones.~\cite{zhou2017selective} use RNNs with a selective gate network that filters the representations based on the first-level representation of the encoder before passing the information to the decoder.~\cite{choi2019event} employ a pointer generator network with both event and global attention. The event attention allows to focus on tokens that describe some event, which is especially important in the news domain, while the global attention captures the general meaning of the sentence. Then, event attention and global attention are summed together as a weighted average. ~\cite{miao2016language} propose a deep generative model with a variational auto-encoder for both abstractive and extractive settings. First, a latent compression is extracted through the generative model, and the variational auto-encoder is used to generate the final compressed sentence according to this compression.

Other RNNs approaches exploit additional features such as contextual embeddings~\cite{zhou2019}, knowledge graphs~\cite{guan2021knowledge} or features from other datasets. For example,~\cite{klerke2016improving} use a three-layer bidirectional LSTM and incorporate an eye-tracking corpus to improve compressions with a multi-task objective.~\cite{Kuvshinova2019Improving} extend the model presented in~\cite{klerke2016improving} by modifying its loss function with the probability score of a generative pre-trained language model.~\cite{guan2021knowledge} performs abstractive sentence compression with knowledge graphs and LSTMs with a multi-copy attention mechanism. In addition,~\cite{zhou2019} propose an unsupervised technique that employs ELMo contextual embeddings with a summarization language model to generate both abstractive and extractive compressions.

Another category of RNN techniques consists of unsupervised techniques. More specifically,~\cite{Malireddy2020} introduce an unsupervised neural compression method that uses an encoder-decoder bi-LSTM architecture with a compressor and a reconstructor. The compressor decides which tokens from the input should be dropped based on a linkage loss that penalizes the model if the reconstructor fails to re-generate the sentence.~\cite{Baziotis2019} use an unsupervised approach with a compressor and a reconstructor similar to~\cite{Malireddy2020}, but instead of optimizing the network for dropping tokens from the input, the reconstructor attempts to rephrase the compressed words. Similarly,~\cite{fevry2018unsupervised} employ a denoising auto-encoder RNN to reconstruct the input text. Such models~\cite{Malireddy2020} have also been used in text summarization, where a pretrained compressor is applied to extracted sentences to improve conciseness in multi-document summarization~\cite{chali2024transfer}.

Even though DL techniques are inherently capable of handling more complex tasks such as abstractive generation, they have been also used for developing extractive sentence compressors. For example,~\cite{nguyen2020contextual} enhance a bi-LSTM model with contextual embeddings for deletion-based compression.~\cite{soares2020sentence} demonstrate that employing syntactic features such as POS tags and dependencies with LSTMs can lead to comparable performance across different LSTM architectures, even when using an order of magnitude less data.  

Other recurrent architectures involve the use of Gated Recurrent Units (GRUs). More specifically,~\cite{Mallinson2018} approach compression using neural machine translation by translating the input sentence into another language and then reverse-translating the output back into the original language. GRUs are also employed in~\cite{nayeem2019diverse} where an abstractive bi-GRU encoder and a simple GRU decoder are combined with a copying attention mechanism. At the same time, to avoid overly compressed sentences, multiple compressions with different compression rates are generated during inference using a diverse beam search decoding strategy.

{\bf CNNs} initialized with BERT embeddings were employed for extractive sentence compression in~\cite{hou2020cnn}.
Even though the experimental results demonstrate that RNNs have stronger performance, this study shows that CNN-based methods achieve a much faster inference time compared to RNN-based ones.

{\bf GNNs} have also been proposed for the task of sentence compression, either in an abstractive~\cite{guan2021integrating} or extractive setting~\cite{park2021sentence}. More specifically,~\cite{park2021sentence} represent each word's dependency tree with GNNs to obtain a richer structural vector representation for each word, while~\cite{guan2021integrating} exploit relations and structural information with GNNs as extracted by FrameNet~\cite{fillmore2001building}. FrameNet is a semantic database that contains approximately 300 lexical items and their semantic relations, as well as their underlying semantic structural dependencies. \cite{zhang2020dense} adopt a graph-based approach that captures higher-level semantic relations with GCN layers and generates abstractive compressions using an LSTM decoder. Finally, SWSum~\cite{guan2025structure} extends this line of research by constructing both a semantic scenario graph and a semantic word-relation graph from FrameNet annotations. Then multiple dual graph-based encoder layers are stacked to iteratively exchange information between the two graphs. The resulting fused graph representations are fed into a transformer decoder to generate compressed abstractive summaries.

{\bf Transformers} have emerged as a powerful model architecture for many NLP tasks, including sentence compression. The performance of Transformer-based architectures in various text generation tasks surpasses that of traditional RNN architectures. Recently,~\cite{west2019bottlesum} apply the principle of Information Bottleneck to both abstractive and extractive sentence compression. The Information Bottleneck method refers to the best trade-off between compression and information loss when compressing a variable $X$ according to relevant information $Y$. In the case of sentence summarization, the variable $X$ represents the generated compression, while the relevant information $Y$ indicates the next sentence of $X$. A pre-trained language model is used to generate the optimal compression for predicting the next sentence. The generated predictions are then used for training a self-supervised abstractive sentence compression model. 

Supervised approaches include the original Transformers architecture~\cite{Vaswani2017} as well as BERT-based models. For example,~\cite{duan2019contrastive} employ a contrastive attention mechanism that considers less relevant parts of the input along with the traditional attention mechanism of Transformers architecture. A softmin function ensures that irrelevant parts of the input will be discouraged.~\cite{Guan2021Frame} use a BERT-based model for abstractive sentence compression that exploits the external knowledge from the FrameNet semantic database. B-BOC~\cite{Shichel2021} is an optimization technique that can be used with any sentence compression model (both supervised and unsupervised) to extract the best candidate sentences for compression. This technique can be used for abstractive sentence compression as well. \cite{guan2023extract} propose a modular abstractive summarization framework which consists of three core components: a) knowledge extraction, where the initial sentence is converted into relation triples, b) content selection, where a RoBERTa-based classifier identifies which triples should be included in the summary and c) rewriting, where a BART-large model re-writes the selected triples into a concise, natural language summary.

Transformers are also used within the framework of knowledge distillation for transferring knowledge from a larger model, which is also known as the teacher, to a smaller model, known as the student. More specifically,~\cite{SclarReferee2022} employ a large pre-trained language model (GPT-3) to distill its knowledge to a GPT-2 model through an iterative distillation process without supervision, where in each round the student acts as a teacher. The generated examples are filtered based on different aspects, including fidelity (whether the compression can be inferred from the input), compression rate, and contextual filter (whether the next sentence could be predicted from the output). The smaller model is then fine-tuned on the final compression corpus as generated by the teacher model. \cite{jung2024impossible} show that a smaller Transformer can serve as its own teacher by generating multiple candidate summaries and filtering them based on their quality, developing a distilled summarization model without relying on a LLM.

\subsubsection{Reinforcement Learning}
Reinforcement learning (RL) has been studied extensively within the wider field of text summarization. It is based on the interaction of the agent with the environment through rewards. The objective of the agent is to learn and apply an action policy (in this setting, e.g., deleting tokens) that maximizes the sum of rewards received over time, either from the environment (unsupervised) or from a labeled dataset (supervised). Since RL does not require a set of labeled data, it is typically met in unsupervised~\cite{Liangguo2018,Ghalandari2022EfficientLearning, hyun2022generating} and semi-supervised~\cite{Zhao2018} settings rather than supervised~\cite{Zhang2021Fact} ones.

The first work to use RL in sentence compression, proposed in~\cite{Liangguo2018}, is based on the Deep Q Network (DQN). DQN is a value-based deep reinforcement learning method that uses neural networks to approximate the expected cumulative reward of an action, under the assumption of following the optimal policy thereafter. This work introduces a DQN-based extractive sentence compression method that incorporates hard constraints from previous IPL approaches~\cite{Clarke2008GlobalApproach} to be considered in the reward function. These constraints include grammatical, syntactic, and length rules. The agent iteratively deletes tokens from the input sentence until the termination of the process. Similar to~\cite{Liangguo2018}, SCRL~\cite{Ghalandari2022EfficientLearning} is another unsupervised extractive approach, starting from a pretrained transformer encoder, which is fine-tuned for sequence classification via reinforcement learning (policy gradient). An interesting aspect is that it is configurable via reward functions (fluency, similarity to source, compression ratio) that can be tailored to specific use cases. Fluency is computed with a masked language model as the average likelihood (logit) of each token in the generated compression.

~\cite{hyun2022generating} use an RL model formulated as a Markov decision process to solve abstractive sentence compression. The agent's rewards include the similarity between the source and the target compression as well as the fluency of the output. The fluency of the generation is computed based on the perplexity of a language model (GPT-2). Low perplexity values indicate high fluency in the output, while higher values indicate low fluency. The backbone of this method is a language model that is pre-trained based on prompt-based text with word-level perturbations and length prompts. The perturbations include shuffles and adding or removing words from the input prompt. The model is trained to reconstruct the initial prompt. At the same time, they employ a controllable multi-summary learning mechanism that is capable of generating multiple summaries of different lengths during inference.

~\cite{Zhao2018} use a bi-directional RNN as a policy-gradient network to decide which tokens should be retained to maximize a reward function. The output of the policy network is passed to a syntax-based language model, which serves as an evaluator to assess the readability and the grammatical coherence of the compression. The readability score of the evaluator acts as a reward to the policy network in order to update its weights accordingly. Their policy is not initialized randomly but with pseudo-compressions yielded by unsupervised models.

FAR-ASS~\cite{Zhang2021Fact} exploits RL in a supervised manner to improve the factual consistency of the generated compressions. The model is trained with a mixed-objective reward function that combines both factual accuracy and the longest common subsequence between the generated and target compressions.

\subsubsection{Summary}
A more detailed analysis regarding the type as well as the training style of each compression method is shown in Table~\ref{tab:methods_type_clustering}. While extractive and supervised approaches seem to dominate the field with more than 67\% and 82\% respectively, the rise of neural models allows for more complex approaches such as abstractive methods with different training styles, including unsupervised, semi-supervised, or weakly supervised, as well as self-supervised techniques with recent Transformer-based architectures.

\subsection{Datasets}
This section presents the most common datasets that are used in extractive and abstractive sentence compression. Some of them are used mainly for evaluation due to their limited size while larger ones are used for both training and evaluation. Table~\ref{tab:datasets} presents a list of these datasets, along with their type (extractive, abstractive), size, year of introduction, whether they are mainly used for evaluation and tasks employed. This section begins by presenting extractive datasets followed by a discussion on abstractive datasets.

\begin{table*}[hbt]
    \centering
    \caption{Datasets for sentence compression. All datasets concern news-related text. Table is ordered by year.}
    \begin{tabular}{c|c|c|c|c}
        Datasets  &  Type  & Size & Year & Evaluation-mainly \\ \hline
        Ziff-Davis~\cite{Knight2000}                     & Extractive  & 1,067 & 2000  &   \\
        \cite{vandeghinste2004sentence} & Extractive & - & 2004 &  \\
        BNC~\cite{Clarke2008GlobalApproach}              &  Extractive  & 1,433 &  2008   & \cmark  \\
        Broadcast~\cite{Clarke2008GlobalApproach}        &  Extractive  & 1,370 & 2008    & \cmark \\
        \cite{yamangil2008mining} & Abstractive & 380K & 2008 &   \\
        Google~\cite{Filippova2013OvercomingCompression} & Extractive  & 250K & 2013  &  \\
        DUC 2004~\cite{Rush2015ASummarization}           &  Abstractive & 500 & 2015   & \cmark \\
        Gigaword~\cite{Napoles2012, Rush2015ASummarization}   &  Abstractive & 3.8M & 2015   &  \\
        MSR~\cite{toutanova-etal-2016-dataset}        &  Abstractive & 26K & 2016 & \\
        Newsroom~\cite{Grusky2018Newsroom:Strategies, Ghalandari2022EfficientLearning}  &  Extractive  &  280 & 2022  &   \cmark \\
    \end{tabular}
    \label{tab:datasets}
\end{table*}

\begin{table}[H]
    \centering
        \caption{Overview of sentence compression methods based on type (extractive, abstractive) and training style (supervised, semi-supervised, weakly-supervised, self-supervised, and unsupervised).}
    \resizebox{\linewidth}{!}{%
    \begin{tabular}{|p{7cm}|c|c||c|c|c|c|c|} \hline
    Method & \multicolumn{2}{c||}{Type} & \multicolumn{5}{c|}{Training Style} \\ \hline
           & \begin{turn}{60}Extractive\end{turn} & \begin{turn}{60}Abstractive\end{turn} & 
           \begin{turn}{60}Supervised\end{turn} &
           \begin{turn}{60}Semi-Supervised\end{turn}  & \begin{turn}{60}Weakly-Supervised\end{turn}  & \begin{turn}{60}Self-Supervised \end{turn} 
           &            \begin{turn}{60}Unsupervised\end{turn} \\ \hline
       \cite{Knight2000, Jing2000, Knight2002, oguro2002evaluation, lin2003improving, vandeghinste2004using, vandeghinste2004sentence, McDonald2006, unno2006trimming, Galley2007, nomoto2007, clarke2007modelling, wu2007, yamangil2008mining, Clarke2008GlobalApproach,zajic2008, hirao2009syntax, Galanis2010, wang2012two, harashima2012flexible, zhang2012research, yoshikawa2012-sentence, thadani2013-sentence, zhang2013chinese, Filippova2013OvercomingCompression, li2013document, thadani2014approximation, Filippova2015, klerke2016improving, Liu2017Attention, wang2017syntax, hasegawa2017japanese, kamigaito2018higher, frederico2018multilingual, Kuvshinova2019Improving, hou2020cnn, nobrega2020sentence, nguyen2020contextual, park2021sentence, zi2021som, kamigaito2020syntactically, soares2020sentence, tavakoli2025sentence} & \cmark &  & \cmark & & & &  \\ \hline
       \cite{Filippova2008, xu2009parse, cordeiro2009unsupervised, fevry2018unsupervised, wang2008automatic, Schumann2020DiscreteExtraction, Malireddy2020, Ghalandari2022EfficientLearning, liu2022learning} & 
       \cmark & &  & & & &  \cmark \\ \hline
       \cite{turner2005supervised} & \cmark & &  & \cmark &  & & \cmark  \\ \hline
       \cite{clarke2006models} & \cmark &  & \cmark &  & \cmark & &\\ \hline
       \cite{Zhao2018}         & \cmark & & & \cmark & &  &  \\ \hline
       \cite{Shichel2021}  & \cmark & & \cmark &  &  & & \cmark\\ \hline
       \cite{juseon2024instructcmp} & \cmark & \cmark & \cmark &  & & & \cmark \\ \hline 
     \cite{Cohn2008, Cohn2009, napoles2011paraphrastic, Galanis2011ACompressor, aziz2012cross, cohn2013, Rush2015ASummarization, Chopra2016, zhou2017selective, li2018ensure, Yu2018operation, Mallinson2018, duan2019contrastive, choi2019event, nayeem2019diverse, duan2019zero, li2020keywords, zheng2020controllable, zhang2020dense, Zhang2021Fact, guan2021knowledge, guan2021integrating, Guan2021Frame, guan2023extract, Lin2023Multimodal, yuan2024exploring, guan2025structure, tollef2025cross} & & \cmark & \cmark & & & & \\ \hline
     \cite{jing2023vision} & & \cmark & \cmark & & & \cmark &  \\ \hline
       \cite{li2020explicit} & & \cmark & \cmark & \cmark & & & \cmark \\ \hline
       \cite{Liu2022Controlling} & & \cmark & \cmark & & & &  \cmark\\ \hline 
    
       \cite{Baziotis2019, zhou2019, SclarReferee2022} & & \cmark & & & & & \cmark\\ \hline
       
       \cite{miao2016language} & \cmark & \cmark & \cmark &  \cmark & & &  \\ \hline
       \cite{west2019bottlesum} & \cmark & \cmark  &  & & & \cmark  & \cmark \\ \hline
    \cite{jung2024impossible} &  & \cmark & & & & \cmark & \\ \hline
    \end{tabular}
}
    \label{tab:methods_type_clustering}
\end{table}

\subsubsection{Extractive Datasets}
Extractive sentence compression datasets consist of long sentences accompanied by their corresponding compression. The compression is obtained by deleting words from the input sentence; thus, it is always a sub-sequence of the original long sentence. It is worth noting that the majority of sentence compression works are limited only to the news domain. 

The first dataset for extractive sentence compression is extracted from the Ziff-Davis corpus~\cite{Knight2000}. Ziff-Davis has a relatively small corpus of newspaper articles announcing computer products. A small portion of the corpus has about 1,000 sentence-compression pairs~\cite{Knight2002}. This dataset has been mostly used for training and evaluating early works on sentence compression. In addition, ~\cite{vandeghinste2004using} build a parallel corpus for subtitle generation that contains transcriptions from several broadcasts. The British National Corpus (BNC) and the English Broadcast News Corpus are two other small human-annotated extractive sentence compression datasets that are typically used for evaluation purposes~\cite{Clarke2008GlobalApproach}. The first one comprises 82 newspaper articles (1,433 sentences) from the BNC and the American News Text corpus, while the second one comprises 50 stories (1,370 sentences) from the HUB-4 1996 English Broadcast News corpus. 

Summarization datasets have also been used for the evaluation of sentence compression. For example, a small extractive dataset of 280 examples is constructed in~\cite{Ghalandari2022EfficientLearning} based on Newsroom. Newsroom~\cite{Grusky2018Newsroom:Strategies} is a large dataset for training and evaluating summarization systems. It contains 1.3 million articles and summaries written by authors and editors in the newsrooms of 38 major publications. The summaries are obtained from search and social metadata between 1998 and 2017 and use a variety of summarization strategies combining extraction and abstraction.~\cite{Ghalandari2022EfficientLearning} extract the first three sentences from each article with a length between 15 and 60 tokens. They consider cases where the tokenized summary was contained in a tokenized sentence. They further filter these cases based on grammaticality and informativeness.

Google sentence compression~\cite{Filippova2013OvercomingCompression} is the largest dataset for extractive sentence compression. It is automatically constructed, consisting of 250K first sentences of a news article along with their extractive compression derived from information in the corresponding article's headline\footnote{\href{https://github.com/google-research-datasets/sentence-compression}{https://github.com/google-research-datasets/sentence-compression}}.

\subsubsection{Abstractive Datasets}
Abstractive datasets contain sentence-compression pairs, where the compression is not necessarily a direct sub-sequence of the original long sentence. While the majority of related literature has focused on the news domain, there also exists work in other domains, such as subtitles or Wikipedia documents. For example,~\cite{yamangil2008mining} exploit Wikipedia revision history to enrich compression data with more than 380,000 compression pairs. 

A small abstractive dataset is constructed in \cite{Rush2015ASummarization} based on the data of Task 1 and 2 of the 2004 Document Understanding Conferences (DUC)\footnote{\href{https://duc.nist.gov/duc2004/}{https://duc.nist.gov/duc2004/}}. It contains 500 news articles from the New York Times and Associated Press Wire services, each paired with four human summaries. A repository containing these data is available\footnote{\href{https://github.com/UsmanNiazi/DUC-2004-Dataset}{https://github.com/UsmanNiazi/DUC-2004-Dataset}}. For the purpose of sentence summarization, the first sentence of each article is used, while summaries are clipped at 75 characters. This leads to summaries with an average of 10 words.

A larger abstractive text compression dataset~\cite{toutanova-etal-2016-dataset} is introduced from Microsoft Research (MSR). It consists of 26,000 long sentences accompanied by their respective compressions. Each long sentence is paired with multiple target compressions (1 to 5) rated by quality judgments regarding sentence coherence and meaning preservation. Besides the newswire domain, the MSR dataset also includes journals as well as business and technical reports. One of the largest datasets for abstractive sentence compression is Gigaword~\cite{Napoles2012}. Gigaword is used in~\cite{Rush2015ASummarization} to create a dataset by pairing the first sentence of each article with its headline. Gigaword is a large-scale dataset that is used for summarization and headline generation. It contains approximately 9.5 million articles from different news sources. After rule-based filtering of some spurious pairs, about 4 million examples remained.

Abstractive datasets for sentence compression are either used only for evaluation (DUC-2004) while larger datasets such as Gigaword~\cite{Napoles2012} and MSR~\cite{toutanova-etal-2016-dataset} are also used for training and evaluation. 

\subsection{Measures}
This section reviews existing evaluation measures for the task of sentence compression, emphasizing widely used metrics such as compression rate and ROUGE score but also other measures that are based on grammatical and lexical relations.

\subsubsection{Compression Rate}
One of the most widely used metric for evaluating sentence compression methods is Compression Rate~\cite{Knight2000, McDonald2006, Galley2007, Galanis2010, Filippova2015, zhou2019, Zhao2018, Ghalandari2022EfficientLearning}. It is defined as the number of words in the compressed text divided by the number of tokens in the uncompressed text as follows: 

\begin{equation}
\text{Compression Rate} = \frac{\text{\# words in compressed sentence}}{\text{\# words in original sentence}}
\end{equation}

Compression rate can serve as a good indicator of the level of compression in the predicted sentence. Predictions with low compression rates might suggest overly compressed sentences which may result in information loss. On the other hand, predictions with high compression rate might not successfully summarize the input sentence. 

However, compression rate alone cannot fully measure and capture other essential factors that contribute to the quality of a compression such as coherence, meaning preservation or fluency. Thus, compression rate should be used to evaluate compression models in combination of additional metrics such as the ROUGE score. 

\subsubsection{ROUGE score}
The family of ROUGE metrics~\cite{Lin2004Rouge} is one of the most popular measures for text summarization. Since text summarization and sentence compression share many similarities in terms of shortening the initial text into a more concise form, ROUGE is also widely used for sentence compression. ROUGE metrics are based on the n-gram overlaps or the longest common subsequence (LCS) between the target and the prediction and can be used to evaluate the quality of the generated text. ROUGE consists of several metrics such as ROUGE-N (ROUGE-1, ROUGE-2, etc.), ROUGE-L, and ROUGE-S. In sentence compression, ROUGE metrics are computed between the predicted and ground truth compressions. More specifically, ROUGE-1 and ROUGE-2 are used for measuring the overlap of unigrams and bigrams between the predicted and ground truth compression, respectively. ROUGE-L is used for measuring the LCS between the generated and the target compression.

Even though ROUGE is an established evaluation metric for sentence compression,~\cite{Zhang2021Fact} highlight some of its limitations. For example, high ROUGE score might be misleading producing factual errors.~\cite{li2018ensure} reveal that existing encoder-decoder models~\cite{zhou2017selective} that achieve significant performance according to established metrics like ROUGE score might not generate faithful compressions. Another important factor in this context is the informativeness of the generated compressions.~\cite{zheng2020controllable} work towards this direction by proposing an entity-controllable compression model that ensures the presence of the most relevant entities of the input sentence in the compressions. This method works by first extracting all the entities from the input sentence and ranking the most important ones with an entity selector module. Finally, instead of conventional left-to-right decoding, the model decodes the text around the selected entities, ensuring that the resulting compression is both concise and informative.

\subsubsection{Other measures}
Other measures include simple token-overlapping~\cite{Ghalandari2022EfficientLearning}, grammatical relations~\cite{clarke2006models, Filippova2013OvercomingCompression} or even human evaluation~\cite{Filippova2013OvercomingCompression}. In specific,~\cite{clarke2006models} demonstrate that measuring the grammatical relations between the input and the target is a measure that correlates well with human judgments. The F1 score between the functional structures of the sentence and its compression, obtained via Lexical-Functional Grammar parsing, was proposed in \cite{Riezler2003StatisticalGrammar} and used in \cite{Filippova2013OvercomingCompression}. In addition, simple F1 score between the tokens of the predicted and ground truth compression are computed in \cite{Ghalandari2022EfficientLearning}. In \cite{Filippova2013OvercomingCompression}, humans were shown pairs of sentences and their compression and were asked to rate in a scale from 1 to 5 the readability (i.e., grammaticallity, fluency) and the informativeness (i.e., importance, representatives) of the compression. Three ratings were collected for every compression.

\subsection{Comparative Evaluation}
This section presents a comparative evaluation of different methods on both abstractive and extractive datasets. Due to the high diversity of the evaluation as well as the large number of publications in the field, we emphasize the most recent works. However, to ensure a fair comparison and analysis, we include some important methods from earlier approaches as well. 

\begin{table}[hbt]
    \caption{Results of sentence compression methods on different extractive datasets. We report ROUGE-1 (R-1), ROUGE-2 (R-2) and ROUGE-L (R-L) scores, F-measure, relations F-measure (Relations F1) and compression rate (CR). All scores are reported in \%. Results are obtained from the corresponding papers. Dashes (-) indicate that the score is not provided.}
    \label{tab:extractive_compression_scores}
    \centering
    \begin{tabular}{c|c|cccccc}
    Dataset     & Method      & \multicolumn{6}{c}{Scores}    \\ \hline
                &             & R-1 & R-2 & R-L & F-measure & Relations F1 & CR \\ \hline
    \multirow{4}{*}{Ziff-Davis} & \cite{Knight2000}    & - & - & - & 68.80 & - & \textbf{70.19}\\
                                & \cite{unno2006trimming}     & - & - & -  & \textbf{78.58} & - & 58.14 \\ 
                                & \cite{Clarke2008GlobalApproach} & - & - & - & - & 46.77 & 48.67 \\                               
                                & \cite{Cohn2009}  & - & - & - & - & \textbf{56.55} & 67.45 \\ \hline  
                                
    \multirow{4}{*}{Google}     
                                & \cite{Filippova2015}  & - & - & - & 82.0  & - & 38.0  \\
                                & \cite{Zhao2018}       & - & - & - & 85.1  & - & 39.0 \\ 
                                & \cite{kamigaito2020syntactically} & \textbf{79.3} & \textbf{71.4} & \textbf{79.1} & 85.5 & - & - \\ 
                                & \cite{park2021sentence} & - & - & - & \textbf{86.2} & - & \textbf{40.9} \\
\hline
    \multirow{2}{*}[-1.3em]{BNC}        & \cite{Clarke2008GlobalApproach}  & - & - & - & -     & \textbf{50.1} & 68.5 \\ 
                                & \cite{thadani2013-sentence}    & - & - & - & 71.91 &   -  & - \\   
                                & \cite{thadani2014approximation}  & - & - & - & 74.00 &   -  & - \\                           
                                & \cite{Ghalandari2022EfficientLearning} & \textbf{79.49} & \textbf{62.32} & \textbf{78.63 }& \textbf{76.50} & - & \textbf{76.0} \\        \hline
                                
    \multirow{2}{*}[-1.3em]{Broadcast}  & \cite{Clarke2008GlobalApproach} & - & - & & - & \textbf{40.8} & 63.7 \\ 
                                & \cite{Ghalandari2022EfficientLearning}  & \textbf{82.22} & \textbf{66.01} & \textbf{81.78} & 78.7 & - & \textbf{78.0} \\
                                & \cite{thadani2013-sentence}             & -     & -     & -     & 77.82 & - & - \\
                                & \cite{thadani2014approximation}         & -     & -     & -     & \textbf{80.13} & - & - \\
 \hline \hline
    \end{tabular}

\end{table}

Table~\ref{tab:extractive_compression_scores} shows the results of different methods on {\em extractive} datasets, where statistical~\cite{Knight2000, unno2006trimming}, ILP~\cite{Clarke2008GlobalApproach, thadani2013-sentence, thadani2014approximation} and dependency parsing~\cite{Cohn2009} approaches as well as neural models such as RNNs~\cite{Filippova2015, Zhao2018, kamigaito2020syntactically} and Transformers~\cite{park2021sentence, Ghalandari2022EfficientLearning} are reported. 

In specific, for the Ziff-Davis corpus, the highest relations F1 (56.55\%) is obtained with a dependency tree that prunes unnecessary words from the input~\cite{Cohn2009}. The highest F-measure (78.58\%) is achieved with a noisy channel model~\cite{unno2006trimming} that is extended with a maximum entropy model. Note that approaches beyond 2013 for the Ziff-Davis corpus are not included in Table~\ref{tab:extractive_compression_scores} as recent methods typically employ larger datasets for evaluating their performance. 

For the Google sentence compression dataset, we can infer from Table~\ref{tab:extractive_compression_scores} that the incorporation of syntactic and linguistic features boosts the performance of the different neural models~\cite{kamigaito2020syntactically, park2021sentence} compared to the baseline LSTM model~\cite{Filippova2015}. The combination of dependency parsing techniques with a Transformer-base model~\cite{park2021sentence} yields the best results with more than 86~\% in terms of F1 score. In addition, the introduction of RL~\cite{Zhao2018} also improves the performance of the baseline model (85.1\% compared to 82\% in terms of F1 score).

BNC and Broadcast datasets have been extensively used for the evaluation of ILP methods. For the Broadcast dataset, an ILP method with Lagrangian relaxation~\cite{thadani2014approximation} shows additional gains against traditional ILP methods with global constraints~\cite{Clarke2008GlobalApproach} or commodity flow constraints~\cite{thadani2013-sentence}. On the other hand, for the BNC dataset, an unsupervised Transformer-based method~\cite{Ghalandari2022EfficientLearning} that employs RL achieves the highest F1 score of all the ILP methods. In addition, a higher compression (76\% compared to 68.5\%) is noticed, which might indicate reduced information loss during the generation of the compression.

\begin{table}[hbt]
    \caption{Results of sentence compression methods on different abstractive datasets. We report ROUGE-1 (R-1), ROUGE-2 (R-2) and ROUGE-L (R-L) scores and compression rate (CR). All scores are reported in \%. The first section of each dataset refers to supervised methods, while the second section, refers to unsupervised methods. Results are obtained from the corresponding papers. Dashes (-) indicate that the score is not provided.}
    \label{tab:abstractive_compression_scores}
    \small
    \centering
    \begin{tabular}{c|c|cccc}
    Dataset     & Method      & \multicolumn{4}{c}{Scores}    \\ \hline
                &             & R-1 & R-2 & R-L & CR \\ \hline
    \multirow{6}{*}[-1.4em]{Gigaword}   & \cite{Rush2015ASummarization}  & 31.00  & 12.65  & 28.34 & -  \\
                                & \cite{Chopra2016}    & 33.78  & 15.97  & 31.15  & -  \\ 
                                & \cite{zheng2020controllable} & 40.03 & 17.93 & 36.74 &  - \\ 
                                & \cite{guan2021knowledge} & \textbf{44.42} & \textbf{31.07} & \textbf{40.71}  & -  \\
                                \cmidrule{2-6}
                                & \cite{Baziotis2019} & 25.39 & 8.21 & 22.68  & - \\
                                & \cite{zhou2019}      & 26.48 & \textbf{10.05} & 24.41  & -  \\
                                & \cite{Malireddy2020} & \textbf{29.80} & 7.52 & 26.10  & - \\
                                & \cite{Ghalandari2022EfficientLearning}  & 29.64 & 9.98 & \textbf{26.57} & 28.0 \\ \hline

    \multirow{4}{*}[-4em]{DUC 2004}   & \cite{Rush2015ASummarization}    & 28.18  & 8.49  & 23.81 & - \\
                                & \cite{Chopra2016}    & 28.97  & 8.26  & 24.06 & -    \\ 
                                & \cite{guan2021integrating} & 30.03 & 10.71 & 26.05 & - \\ 
                                & \cite{zheng2020controllable} & \textbf{34.69} & \textbf{14.39} &  \textbf{31.07} & -  \\ 
                                \cmidrule{2-6}
                                & \cite{Baziotis2019}  & 22.13  & 6.18  & 19.3  & - \\
                                & \cite{Malireddy2020} & 22.92  & 5.52  & 19.85  & -  \\
                                & \cite{Ghalandari2022EfficientLearning} & 25.27 & 7.82 & 22.14 & 35.0 \\ 
                                & \cite{Schumann2020DiscreteExtraction} & 26.04 & 8.06 & 22.90 & -\\ 
                                & \cite{liu2022learning} & 26.71 &  7.68 & 23.06 & -  \\  
                                & \cite{hyun2022generating} & \textbf{27.88} &\textbf{ 9.35} & \textbf{24.49} & - \\
                                \hline
                                
\end{tabular}
\end{table}

Table~\ref{tab:abstractive_compression_scores} shows the results of different methods on {\em abstractive} sentence compression datasets, where different neural models are reported. Among all the supervised methods for the Gigaword dataset, the feed-forward network with a convolutional attention-based encoder~\cite{Rush2015ASummarization} achieves a moderate score of 31\% in terms of ROUGE-1. Even though this method outperforms all the unsupervised approaches~\cite{Baziotis2019, zhou2019, Malireddy2020, Ghalandari2022EfficientLearning}, it fails to compete with the supervised RNN-based methods~\cite{Chopra2016, zheng2020controllable, guan2021knowledge}. More specifically, the attention-based RNN~\cite{Chopra2016} demonstrates a slight improvement with 33.78\% ROUGE-1 which indicates that indeed RNNs are more powerful in this context, allowing for capturing longer dependencies and as a result generating more relevant compressions. An important boost in performance is noticed when LSTMs are paired with controllable attributes such as guiding entities~\cite{zheng2020controllable}, resulting in an increase of approximately 7\% points in terms of the ROUGE-1 score. The same conclusions are drawn with the incorporation of knowledge graphs~\cite{guan2021knowledge} with a significant increase in performance of 44.42\% and 31.07\% in terms of ROUGE-1 and ROUGE-2 scores, respectively. Regarding the unsupervised approaches for the Gigaword dataset, the LSTM-based method with the compressor and reconstructor~\cite{Baziotis2019} achieves a relatively low ROUGE-1 score of 25.39\%, while slight improvements are noticed when ELMo embeddings are added in a similar LSTM architecture~\cite{zhou2019}. The approach of~\cite{Malireddy2020}, which optimizes the LSTM architecture to not only reconstruct but also rephrase words within the text, demonstrates additional gains with a noticeable improvement of 29.80\% ROUGE-1 score. In addition, Transformers with RL~\cite{Ghalandari2022EfficientLearning} achieve a very similar ROUGE-1 score (29.64\%) but a slightly higher ROUGE-L score (26.57\%). This might indicate that while Transformers are powerful in many NLP tasks, RNN architectures are still useful in this context.

Similar trends are observed on the DUC 2004 dataset, where the same supervised and unsupervised techniques are reported. The supervised approaches, such as the feed-forward neural network~\cite{Rush2015ASummarization}, attention-based RNN~\cite{Chopra2016} and the LSTM with knowledge graphs~\cite{guan2021knowledge} demonstrate comparable performance with approximately $30\%$ ROUGE-1 score, while the LSTM with guiding entities outperforms all the previous approaches with more than 34\% ROUGE-1 score. Among the unsupervised approaches, we notice the same patterns with the Gigaword dataset, while there are slight improvements specifically with the incorporation of discrete optimization techniques~\cite{Schumann2020DiscreteExtraction, liu2022learning} and the use of RL with Transformers~\cite{hyun2022generating}. Since DUC 2004 is typically used only for evaluation, the ROUGE scores for both the unsupervised and supervised methods remain relatively low compared to the Gigaword dataset. Similar to extractive models in Table~\ref{tab:extractive_compression_scores}, we notice again that the incorporation of additional features and techniques into sequence-to-sequence models can improve the overall performance of compression models.

The overall comparative evaluation demonstrates that methods that employ Transformer-based architectures~\cite{hyun2022generating, Ghalandari2022EfficientLearning} seems to outperform the majority of existing methods for both supervised and unsupervised models. On the other hand, in some cases, LSTM architectures can compete with Transformers, achieving similar ROUGE scores in an unsupervised setting. Additional gains are observed when Transformers are combined with other techniques such as RL or controllable attributes i.e., length and entities. These results also indicate the potential benefits of different attribute controls in generating more accurate and relevant compressions. Finally, we notice that methods on extractive datasets achieve overall higher ROUGE scores than methods on abstractive datasets. Extractive methods have a significant advantage over ROUGE scores since they are based on deletion-only operations, in contrast to abstractive methods, which are based on rewriting or paraphrasing operations.

\subsection{Other Trends}
In this subsection, we explore additional trends within the literature of sentence compression such as methods employed for length control as well as approaches related to multilingual and multimodal sentence compression.

\subsubsection{Length Control in Sentence Compression}
Length-controllable approaches for sentence summarization~\cite{Baziotis2019, zhou2019} are attracting increasing attention during the last years. Several studies incorporate an implicit length control to improve the quality of the generated compressions, while others control the length explicitly during inference. For example, \cite{zhou2019} applies a length penalty to avoid very short compressions,  while explicit approaches use decoder embeddings~\cite{fevry2018unsupervised, Baziotis2019}, discrete optimization search~\cite{Schumann2020DiscreteExtraction, liu2022learning} or dynamic programming~\cite{Liu2022Controlling}. This subsection discusses explicit control in sentence summarization.

Explicit length-controllable models can be either character-level~\cite{napoles2011paraphrastic, Liu2022Controlling, SclarReferee2022} or word-level~\cite{fevry2018unsupervised, Baziotis2019, Schumann2020DiscreteExtraction, liu2022learning, juseon2024instructcmp}. The first abstractive work to perform length control at the character level is introduced in~\cite{napoles2011paraphrastic}, where the length control is achieved based on dynamic programming to find the optimal number of paraphrases for reducing the character length of the compression. Recently, another character-length control approach was proposed in~\cite{Liu2022Controlling}, where the length control is formalized as a 0/1 knapsack problem, where the value is formulated as the predicted word's log-likelihood and the weight is represented as the number of characters in the word. For more efficient inference, different lengths are divided into buckets. This method can be either supervised (using existing sentence compression datasets) or unsupervised (using pseudo-compressions generated by~\cite{Schumann2020DiscreteExtraction}). Finally, REFEREE~\cite{SclarReferee2022} is a sentence compression method that is based on symbolic knowledge distillation~\cite{west2022symbolic}, where knowledge is transferred from a large language model to a significantly smaller one. REFEREE is composed of two models: a distillation model (REFEREE-distill) which transfers the knowledge to the smaller model, and a length-controllable model (REFEREE-control) which is trained on the generated examples from the distillation process. 

Despite some efforts for character-level length control, the majority of the existing literature focuses on word-level length control in an unsupervised setting. More specifically,~\cite{Schumann2020DiscreteExtraction} propose an unsupervised length-controllable method for extractive sentence compression that is based on a discrete search optimization strategy. The length control is performed at the word level by extracting source words from the input sentence based on hard length constraints. The best sentence compression is selected based on a first-choice hill-climbing~\cite{raffel17a} approach according to a hand-crafted heuristic function that aims to maximize predefined aspects such as fluency and similarity to the source. This approach can explicitly control the length of the generated compressions by searching across the space of all the feasible compressions. Despite the high performance of this method, it is slow and computationally exhausting. Another unsupervised word-level length controllable approach is proposed in~\cite{liu2022learning} where dynamic programming decoding is employed to control the length of the output. This approach allows for faster inference compared to~\cite{Schumann2020DiscreteExtraction}, since the exhaustive search is performed only for generating the data for the model's training. Both methods are extractive since they rely only on deletion operations. On the other hand, existing auto-encoder approaches can be either extractive~\cite{fevry2018unsupervised} or abstractive~\cite{Baziotis2019}. The length control is achieved in both approaches with a length embedding added to the decoder to influence the length of the generated compressions. Finally, InstructCMP~\cite{juseon2024instructcmp} introduces a zero-shot length control approach that employs Large Language Models (LLMs), such as  ChatGPT, to restrict the length of the compression. The best results are obtained when the prompt instructions include both the desired target length of the compression and the original sentence's word count, along with the number of words to remove. Finally, a recent line of work introduces explicit length control for cross-lingual compression, where the desired compression ratio is achieved through a token prepended to the input, enabling a single model to generate translations at different target lengths~\cite{tollef2025cross}.

\subsubsection{Multilingual Sentence Compression}
Besides monolingual sentence compression, there have been a few attempts at multilingual~\cite{frederico2018multilingual} and cross-lingual sentence compression~\cite{aziz2012cross, duan2019contrastive, gasan2024multi, tollef2025cross}. In multilingual sentence compression, the goal is to generate compressed versions of a given sentence in two or more different languages, while in cross-lingual sentence compression, the target compression should be in a different language from the input sentence. A multilingual approach for both English and Portuguese was proposed in~\cite{frederico2018multilingual}, where multilingual representations were obtained by mapping monolingual embeddings into a common multilingual vector space. On the other hand, cross-lingual techniques work either by translating the output to the target language~\cite{aziz2012cross, gasan2024multi}, jointly learning translation and compression from compressed parallel corpora~\cite{tollef2025cross}, or performing zero-shot techniques without the need for annotated data~\cite{duan2019zero}. Even though the majority of monolingual works study the task for the English language, there is also literature for Chinese~\cite{xu2009parse, zhao2022simple, zi2021som, zhang2012research, zhang2013chinese}, Japanese~\cite{hasegawa2017japanese, hirao2009syntax, harashima2012flexible, oguro2002evaluation}, Portuguese~\cite{nobrega2020sentence, frederico2018multilingual}, and Persian~\cite{tavakoli2025sentence} as well.

\subsubsection{Multimodal Sentence Compression}
Multimodal sentence summarization (MMSS), a relatively new field in the era of multimodality, refers to the process of generating short and informative summaries by integrating information from multiple data types such as text, images, audio and videos. Unlike traditional sentence summarization, which relies only on text, multimodal summarization combines these sources to provide a summary that reflects information from the different content types.

Existing approaches mainly focus on generating compressions by combining information from a source image and text. For example, BART-MMSS~\cite{Lin2023Multimodal} adapts BART for multimodal sentence summarization by integrating image information using a Vision Transformer, which extracts image embeddings and injects them into the BART model. In addition, it employs an explicit critical token learning module to identify important tokens in the source sentence, aligning them with visual information in the image, to ensure that they will be included in the final summary. 

Another approach that employs generative pre-trained models for MMSS is Vision-GPLM~\cite{jing2023vision}. Vision-GPLM uses two different encoders for text and image features and merges them through multi-head attention. The model is trained in two phases: a pre-training phase, where only the visual encoder is trained, and a fine-tuning phase, where all components are trained jointly for MMSS. 

\cite{yuan2024exploring} introduce a framework based on the Information Bottleneck principle to more effectively manage the incorporation of visual elements. While existing methods often risk including too much (over-preservation) or too little (over-compression) visual detail~\cite{Lin2023Multimodal}, this approach works by creating a maximally compressed mapping of the input data, where only information essential to the summary is retained.

\section{Sentence Splitting}
\label{sec:sentence_splitting}
This section reviews methods, datasets, and evaluation metrics for the task of sentence splitting as well as provides a comparative evaluation of the most representative methods for this task. 

\subsection{Methods}
The importance of sentence splitting lies in its ability to break down complex piece of text into shorter and coherent segments. There have been efforts~\cite{nomoto2022fewer} to analyze whether sentence splitting enhances the readability of a complicated sentence. In this context, annotators are asked to rate two-sentence and three-sentence splits from existing splitting datasets. Then, a Bayesian logistic regression model is used to examine the ratings. This model employs different linguistic and cognitive features, including cohesion, perception, and structure, among others. The primary finding of this study indicates that sentence splitting indeed allows for a better understanding of the text. In addition, readability is found to improve more with a two-sentence split than with additional splits~\cite{nomoto2023does}.

In the majority of existing works, sentence splitting is typically met as an intermediate component of sentence simplification~\cite{liu2025community} or discourse relation systems~\cite{yu2019gumdrop}, rather than being handled as an independent task. For example, as part of the \textit{DISRPT 2019 Shared Task} for identifying discourse relations in text spans,~\cite{yu2019gumdrop} employ a sentence splitter using a stacking ensemble method leveraging the outputs of multiple models including deep learning, machine learning, and rule-based techniques. In addition, sentence splitting is also used as a preprocessing step of machine translation methods~\cite{lee2008transformation}.

Methods for sentence splitting can be syntax-based~\cite{Siddharthan2006, Siddharthan2014}, discourse-based~\cite{Cripwell2021} or semantic-based~\cite{kim-etal-2021-bisect}. Syntax-based methods employ syntax or grammar rules to split the input sentence into shorter ones, while discourse-based methods consider the discourse relations within the input sentence. On the other hand, semantic-based methods typically rely on models trained on existing splitting corpora to learn to split the input sentence according to its semantics. 

\begin{figure}
    \centering
    \includegraphics[width=0.7\textwidth]{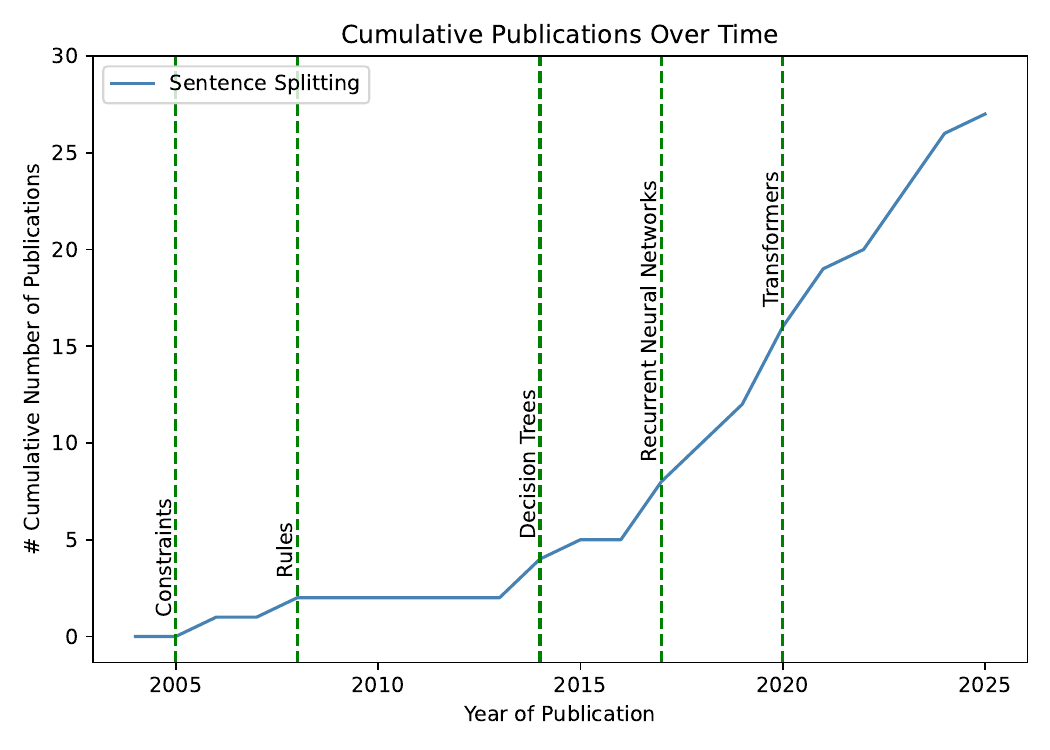}
    \vspace{1.5em}
    \caption{Cumulative number of publications per year from 2005 to 2025 for sentence splitting along with landmarks of key technology adoption.}
    \label{fig:sentence_splitting_cumulative}
\end{figure}

Figure~\ref{fig:sentence_splitting_cumulative} shows the cumulative number of publications from 2005 to 2024 along with landmarks of key technology adoption. A significant increase in interest is evident after 2017 possibly driven by the introduction of DL methods such as RNNs and Transformers. This growth further highlights the importance of sentence splitting for improving various NLP tasks. 

\subsubsection{Syntax-based methods}
First works on sentence splitting appear to be closely related to sentence simplification~\cite{Siddharthan2006, Siddharthan2014}. These works formulate the task of sentence splitting as a constraint satisfaction problem where the goal is to find a valid solution that satisfies a set of constraints. For example,~\cite{Siddharthan2006} creates a set of manually written grammar and syntactic constraints that were then enriched to form a larger set of more extensive ones~\cite{Siddharthan2014}.~\cite{lee2008transformation} start with an initial set of splitting rules and then use a transformation-based learning method to iteratively refine them.~\cite{lee2017splitting} employ a decision tree with different attributes such as sentence length, syntactic dependency patterns and POS labeling. 

\subsubsection{Discourse-based methods}
~\cite{Cripwell2021} introduce a sentence-splitting model based on discourse structure. This method suggests two approaches: a discourse tree model as well a Transformer model. As a result, it falls both into discourse-based and semantic-based categories. This approach ensures that the discourse relation in the original long sentence is preserved in the output of shortened sentences, but requires the presence of a discourse connective, which is not always present when dealing with real sentences. 

\subsubsection{Semantic-based methods}
Recent work addresses the task of sentence splitting with semantic-based methods using supervised training of sequence-to-sequence models on splitting datasets such as WebSplit~\cite{narayan-etal-2017-split, aharoni-goldberg-2018-split}, WikiSplit~\cite{botha-etal-2018-learning} and BiSECT~\cite{kim-etal-2021-bisect}. For example,~\cite{narayan-etal-2017-split} use a three-layered LSTM, while other approaches~\cite{aharoni-goldberg-2018-split, botha-etal-2018-learning, Zhang2020} use a bi-directional attention-based LSTM~\cite{Bahdanau2014} with a copying mechanism similar to a pointer-generator network~\cite{See2017GetNetworks}. In~\cite{wang2020hierarchical} a separator network is used to enhance the standard attention-based LSTM. The separator network determines the content of each split and then forms a representation that is passed to the decoder for generating the split sentence. This method mitigates the repetition issues of previous models that are typically found within the split sentences. Another method to address repetition issues is proposed in~\cite{fan2020memory}, where a memory-based Transformer model is introduced. This method incorporates a memory gate layer positioned between the encoder and the decoder to filter out previously decoded information. In addition, the model is trained with a multi-task learning objective that exploits (Resource Description Framework) RDF triples to encourage the generation of factually correct sentences.

~\cite{berlanga2021split} adopt a cross-lingual approach for both English and Portuguese. They use a GRU model with an attention mechanism that is trained with a symbolic vocabulary. The symbolic vocabulary, which is based on POS tags, allows for generalization in different languages. To transform the generated symbolic tokens into actual text, different versions of MLM BERT models are used. ~\cite{kim-etal-2021-bisect} employ the Transformer architecture to propose several baseline models. More specifically, this works identifies three types of splits according to the volume of modifications: a) direct insertion, b) changes near split, and c) changes across the sentence. Direction insertion involves minimal or no modifications of the initial sentence while changes near split require slight adjustments around the splitting point. Changes across sentence involve significant changes to reconstruct the initial sentence. This works employs a BERT model with an custom loss function that takes into account these split categories.~\cite{guo2020fact} propose a multi-task learning approach to learn sentence splitting and factual correctness. A sequence-to-sequence model is trained not only for sentence splitting but also for determining the correctness of the given facts according to the input sentence. This approach is applied to both LSTM and Transformer models. Another Transformer-based method is introduced in~\cite{alajlouni2023knowledge}, where a split-and-rephrase approach is proposed for transferring knowledge from a rule-based model to a pre-trained language model. First, candidate splits are generated based on the rule-based model, which are then filtered according to meaning preservation, grammatical correctness and simplicity. The filtered samples are used to fine-tune a T5 Transformer model for sentence splitting. Although most research focuses on English, sentence splitting has also been examined in typologically diverse languages such as Persian~\cite{talebpour2024split}, where encoder–decoder models have shown strong performance, demonstrating the broader cross-lingual applicability of the task.

Even though LLMs have achieved great results in various NLP tasks, there is only one work that is focused on their use for sentence splitting~\cite{ponce-etal-2024-split}. More specifically,~\cite{ponce-etal-2024-split} evaluate different LLMs such as Alpaca, GPT-4 Turbo and Pythia, for the split and rephrase task. LLMs were prompted to generate split versions of complex sentences, in different settings, including zero-shot, few-shot in-context learning, and fine-tuning. Regarding fine-tuning, results show that around 3,000 examples of complex and split sentences are needed to achieve reasonable accuracy, with 4,000 to 6,000 pairs to yield the best performance depending on the domain. In addition, instruction-tuned models show promising results with as few as five curated examples.

Table~\ref{tab:splitting_methods_type_clustering} provides a list of splitting methods along with their corresponding category (semantic, discourse, or syntax-based). While earlier approaches mainly rely on syntax-based approaches that formulate the task with constraints~\cite{Siddharthan2006, Siddharthan2014} or employ simple rule-based methods~\cite{lee2008transformation}, the growth of DL methods (RNNs and Transformers) led to the rise of semantic-based approaches~\cite{narayan-etal-2017-split, aharoni-goldberg-2018-split, kim-etal-2021-bisect}.

\begin{table*}[ht]
    \centering
        \caption{Overview of splitting methods based on category (syntax, discourse and semantic-based) and employed approach (constraints, decision trees, discourse trees, RNNs and Transformers). All methods except constraint-based methods are supervised.}
            \resizebox{\textwidth}{!}{%
    \begin{tabular}{|p{4cm}|c|c|c||c|c|c|c|c|c|} \hline
    Paper & \multicolumn{3}{c|}{Category} & \multicolumn{6}{c|}{Approach}  \\ \hline
           & \begin{turn}{60}
               Semantic-based 
           \end{turn}& \begin{turn}{60}Discourse-based\end{turn} & \begin{turn}{60}Syntax-based\end{turn} & \begin{turn}{60}Constraints \end{turn} & \begin{turn}{60}Rules\end{turn}& \begin{turn}{60}Decision Trees\end{turn} & \begin{turn}{60}Discourse Trees\end{turn} & \begin{turn}{60}RNNs\end{turn} & \begin{turn}{60}Transformers\end{turn} \\ \hline
     \cite{narayan-etal-2017-split, aharoni-goldberg-2018-split, botha-etal-2018-learning, Zhang2020, berlanga2021split} & \cmark & & & & & & & \cmark &\\ \hline
     \cite{fan2020memory, kim-etal-2021-bisect, ponce-etal-2024-split, talebpour2024split} & \cmark & & & & & & & & \cmark \\ \hline
    \cite{alajlouni2023knowledge} & \cmark & & & & \cmark & & & & \cmark \\ \hline
    \cite{guo2020fact} & \cmark & & & & & & & \cmark & \cmark \\ \hline
     \cite{Siddharthan2006, Siddharthan2014} & & & \cmark & \cmark & & &  & & \\ \hline
    \cite{lee2008transformation} & & & \cmark & & \cmark   & &  & & \\   \hline
     \cite{lee2017splitting} & & & \cmark & & & \cmark &  & & \\   \hline
     \cite{Cripwell2021} & \cmark & \cmark & &  & & & \cmark & &  \cmark \\ \hline

    \end{tabular}}
    \label{tab:splitting_methods_type_clustering}
\end{table*}

\subsection{Datasets}
The majority of sentence splitting models rely on large annotated splitting datasets. Table~\ref{tab:splitting_datasets} summarizes existing splitting datasets, along with their size and the year of introduction.

WebSplit \cite{narayan-etal-2017-split} is a large-scale synthetic dataset that consists of 5,546 complex sentences extracted by the RDF triples of the WebNLG~\cite{Gardent2017} corpus. The WebNLG corpus consists of RDF triples along with one or more verbalizations, i.e., RDF triples converted to human-readable text. The RDF triples contain relations between the different entities for each text. The process of constructing the WebSplit dataset begins with an initial sentence and its associated RDF triples. A subset of RDF triples for this sentence is identified, and the verbalizations of these RDF triples (shorter sentences) are then merged to create the split version of the original input sentence.

~\cite{Zhang2020} reveal that WebSplit follows easily exploitable syntactic patterns that even simple rule-based techniques can recognize. An improved version of WebSplit is proposed in~\cite{aharoni-goldberg-2018-split} that introduced an alternative training, validation and test set split with less overlap between splits. Still, WebSplit uses a relatively small vocabulary (3,311 words) and contains some unnatural expressions due to its synthetic nature. 

WikiSplit~\cite{botha-etal-2018-learning} contains one million natural English sentence splits that are created automatically from user edits of Wikipedia revision history. WikiSplit's vocabulary is 90 times larger than WebSplit. At the same time it provides more coherent and naturally occurring examples. However, since the dataset is constructed automatically, the preservation of meaning within the examples cannot be guaranteed. As a result, this dataset is typically used to augment training data for enhancing the performance of splitting models. 

Wiki-BM~\cite{Zhang2020} is a subset of the original test set of WikiSplit that contains 500 human-annotated examples out of the total 5,000. MinWikiSplit~\cite{niklaus-etal-2019-minwikisplit} is a large-scale dataset for sentence splitting that builds upon WikiSplit using the principle of minimal prepositions. It contains 203K complex sentences paired with their split sentences according to the principle of minimality, where each sentence is split into a set of minimal propositions. It was originally proposed to improve WikiSplit's split sentences since some sentences could be further split. Each sentence in MinWikiSplit is split into minimally shortened sentences based on human-written, handcrafted rules.

BiSECT~\cite{kim-etal-2021-bisect} contains one million complex sentences accompanied with shorter sentences in four different languages including English, German, French and Spanish. The dataset is created with bilingual pivoting by mapping 1-2 and 2-1 sentence pairs in a bilingual dataset and then translating the pairs into one language. Even though this dataset is typically used for sentence simplification, in this study we focus on the reduction of the sentence length without taking the content into account. 

WikiSplit++~\cite{tsukagoshi2024wikisplit} contains over 600,000 examples of sentence splits from WikiSplit, addressing issues with factual consistency and coherence in the data. Unlike WikiSplit, which may include hallucinations or inaccuracies due to its automated construction, WikiSplit++ uses a natural language inference classifier to filter out unreliable examples, specifically removing those where the simplified sentence does not match the meaning of the original complex sentence. Then, sentence order is reversed in the output to encourage models to generate more diversified outputs rather than simply copying the word sequence of the original complex sentence.

\begin{table}[hbt]
    \centering
    \caption{Datasets for sentence splitting.}

    \begin{tabular}{c|c|c}
        Datasets  & Size & Year \\ \hline
        WebSplit (v.0.1)~\cite{narayan-etal-2017-split}  & 5,546 & 2017  \\
         WebSplit (v.1.0)~\cite{aharoni-goldberg-2018-split}  & 18,830 & 2018  \\
        WikiSplit~\cite{botha-etal-2018-learning} & 999,944  & 2018  \\
        MinWikiSplit~\cite{niklaus-etal-2019-minwikisplit} & 203K & 2019 \\
        Wiki-BM~\cite{Zhang2020}  & 500  & 2020   \\
        BiSECT~\cite{kim-etal-2021-bisect} & 1M & 2021 \\ 
        WikiSplit++~\cite{tsukagoshi2024wikisplit} & 630,433 & 2024 \\
    \end{tabular}
    \label{tab:splitting_datasets}
\end{table}

\subsection{Evaluation Metrics}
The most commonly used metric for assessing the performance of sentence splitting models is BLEU~\cite{Papineni2002}, which is based on n-gram precision. BLEU is used to compute the similarity between a ground truth set of sentences and a predicted set of sentences as follows: 

\begin{equation}
    \textrm{BLEU} = \textrm{BP} \cdot\exp\bigg(\sum_{n=1}^{N} w_{n}\log p_{n}\bigg),
\end{equation}

where $BP$ indicates the brevity penalty used to penalize shorter sequences, $N$ is the maximum n-gram value, $w_{n}$ corresponds to the assigned n-gram weight, which is a value between 0 and 1, and $p_{n}$ is the modified n-gram precision. The baseline value of $w_{n}$ is $\frac{1}{N}$ with $N = 4$. 

The brevity penalty is computed as follows: 
\begin{equation}
BP =\left\{ 
\begin{array}{cl}
1 & \textrm{if } c > r \\
\exp^{(1-r/c)} & \textrm{if } c \leq r
\end{array}\right.
\end{equation}

, where $c$ and $r$ indicate the length of the target and the reference candidate sentence. 

The modified n-gram precision is defined as follows:
\begin{equation}
   p_{n} = 
   \frac{\sum\limits_{C \in \{Candidates\}}\sum\limits_{\textrm{n-gram} \in C}Count_{clip}{(\textrm{n-gram})}}
        {\sum\limits_{C' \in \{Candidates\}}\sum\limits_{\textrm{n-gram'} \in C'}Count(\textrm{n-gram'})}
\end{equation}

, where $Candidates$ represent the set of generated sentences,  $count$ indicates the count of n-grams in the candidate sentence, and $Count_{clip} = min(count, max\_ref\_count)$ with $\max\_ref\_count$ used to limit count to the maximum count of the n-gram in the reference candidate.

While BLEU can serve as an indicator of the quality of the generated text according to the ground truth, it is not capable of taking into account the semantics of a sentence as other metrics, i.e., BERTScore~\cite{Zhang2020BertScore}. BERTScore, which is based on the Transformer architecture, is widely used to measure the quality of generated text in various natural language generation tasks, such as machine translation, summarization, etc. Unlike BLEU, which relies on simple n-gram overlaps, BERTScore exploits BERT's contextual embeddings. To compute the similarity between a ground truth and a predicted set of sentences, the cosine similarities of each token embedding of the input sentences are employed. More specifically, BERTscore calculates the recall and precision for each token embedding in the generated and target sentences. BERTScore then aggregates the F1 scores to provide an overall similarity score that is used to measure how well the generated text reflects the ground truth as follows: 

\begin{equation}
    BERT_{score} = \frac{2\cdot P_{BERT}\cdot R_{BERT}}{ P_{BERT}+ R_{BERT}},
\end{equation}

where $P_{BERT}$ and $R_{BERT}$ indicate the precision and recall between the generated and the reference candidate.



Several works also report the ratio of simple sentences per complex sentence ($\#S/C$), as well as the ratio of tokens per simple sentence ($\#T/S$). The $\#S/C$ ratio can give a rough estimation of the complexity of the given pair of sentences. A higher $\#S/C$  ratio indicates a higher proportion of simple sentences. Similarly, the $\#T/S$ score represents the ratio of the average number of tokens per simple sentence. A low $\#T/S$ ratio indicates that fewer tokens are included in simple sentences while sentences with higher scores tend to contain more tokens. These ratios provide valuable insights into both the complexity and the structure of the generated sentences. Typically, these metrics are computed in relation to the ground truth sentences within a dataset. 


\subsection{Comparative Evaluation}
A comparison across different methods on different datasets for sentence splitting is shown in Table~\ref{tab:splitting_scores}. Similar to sentence compression, we emphasize the most recent methods for sentence splitting. 

\begin{table}[hbt!]
    \caption{Results of splitting methods on different datasets. BLEU, BertScore, $\#S/C$ and $\#T/S$ are reported. Results are obtained from the corresponding papers. Dashes (-) indicate that the score is not provided.}
    \label{tab:splitting_scores}    
    \centering
    \begin{tabular}{c|c|cccc}
     &  & \multicolumn{4}{c}{Scores} \\ \hline
    Dataset & Method & BLEU(\%) & BERTScore(\%) & \#S/C & \#T/S \\ \hline
    \multirow{3}{*}[-2em]{WebSplit(v1.0)}   
    & Reference & - & - & 2.5 & 10.9 \\
    & \cite{aharoni-goldberg-2018-split} & 25.47 & -  & 2.3 & 11.8 \\
                                       & \cite{botha-etal-2018-learning} & 30.5 & - & 2.0 & 8.8 \\    
                                       & \cite{Zhang2020} (LSTM) & 62.6 & - & - & - \\     
                                       & \cite{Zhang2020} (rules)   & 65.9 & - & - & - \\
                                       & \cite{fan2020memory} & \textbf{71.2} & - & 2.5 & 10.3 \\
                                       \hline
                                    
    \multirow{3}{*}[0.5em]{Wiki-BM}    & \cite{ponce-etal-2024-split} & 74.40 & - & -  & - \\  
                                       & \cite{Zhang2020} (rules)   & 77.2 & - & - & - \\ 
                                       & \cite{Zhang2020} (LSTM) & \textbf{87.0} & - & - & -\\ 
                                       \hline
    \multirow{2}{*}{BiSECT}  
    & \cite{ponce-etal-2024-split} & 42.4 &  - & - & - \\
    & \cite{kim-etal-2021-bisect} & 45.8 & \textbf{85.6} & - & - \\
                                       & \cite{aharoni-goldberg-2018-split} & \textbf{46.3} & 85.3 & - & -\\ \hline
    \end{tabular}
\end{table}

Regarding the WebSplit dataset, we notice that the use of simple rules~\cite{Zhang2020} outperforms all the RNN-based models with a BLEU score of 65.9\% compared to 62.6\% of the best performing LSTM model. This finding can be attributed to the simple and repetitive patterns of WebSplit as highlighted in~\cite{Zhang2020}. On the other hand, the Transformer memory-based model of~\cite{fan2020memory} achieves the best results in terms of BLEU score and the closest $\#S/C$ and $\#T/C$ ratios to the reference ones. 

Regarding the Wiki-BM test set, the best performance is obtained using an LSTM with attention and a copying mechanism~\cite{Zhang2020}. Note that this model is trained on the WikiSplit dataset. Unlike WebSplit, the rule-based method does not perform as well on Wiki-BM, since it is a more challenging dataset with more complex examples. 

For the BiSECT corpus, both the RNN-based model~\cite{aharoni-goldberg-2018-split} and the Transformer-based model~\cite{kim-etal-2021-bisect} achieve comparable results ($\sim46\%$ in terms of BLEU and $\sim85\%$ in terms of BERTScore).
  
In conclusion, the overall comparative evaluation reveals an interesting trend regarding the rule-based methods when dealing with simpler datasets such as WebSplit. On the other hand, these methods struggle to perform when more complex datasets such as Wiki-BM are introduced. Transformer and LSTM models consistently achieve the best performance across the different splitting datasets. In addition, LLMs did not outperform traditional neural models in either the Wiki-BM or BiSECT test sets. Interestingly, this limitation suggests a promising research direction for further enhancing LLMs to better handle these tasks.

\section{Challenges and Limitations}
\label{sec:discussion}
This section provides an overview of the challenges and limitations of current methods for addressing the issue of long sentences.

\subsection{Sentence Compression vs Sentence Splitting}
Sentence compression and sentence splitting can both be used to simplify long sentences. Sentence compression can effectively reduce the length of the input sentence, making it more clear and concise. However, this process inherently results in information loss since significant details of the input sentence often get removed. Some sentences can be compressed without losing important information, while others may entail the removal of critical content. On the other hand, sentence splitting usually does not result in information loss, while achieves at the same time both meaning preservation and length reduction for each sentence. However, sentence splitting may not always be a feasible solution, especially when dealing with relatively short sentences. As a result, the choice between sentence compression and sentence splitting often depends on the specific context at hand. The optimal balance between these techniques and the optimization of their use remains an interesting and ongoing challenge.

\subsection{Evaluation Metrics}
Despite the significant progress on both sentence compression and sentence splitting methods, the field lacks an established automated evaluation metric that is capable of measuring the meaning preservation between the generated compression and the original long sentence. While several evaluation metrics exist to evaluate the quality of the resolved sentences, such as ROUGE, BLUE or grammatical F1 score, these are focused only on the overlap of tokens as well as grammatical or syntactic consistency. The evaluation of meaning preservation is typically assessed only by human annotators, which significantly limits its scalability and introduces additional effort and costs. In fact, there is only one work~\cite{Beauchemin2023Meaning} for evaluating meaning preservation within the field of sentence simplification, but currently there are no established metrics for sentence compression and sentence splitting for assessing this critical aspect. The importance of a meaning preservation metric becomes evident when we consider the primary objective of resolving a long sentence. By retaining the original meaning of a sentence while shortening or splitting it, we ensure that the resulting text will remain logically coherent and will deliver accurately the intended message. This is especially important in contexts where preserving the meaning is crucial, such as news articles, social media posts or legal documents. Such a metric can significantly advance research and benefit practical applications in sentence compression and sentence splitting.

\subsection{Interpretability}
RNNs and Transformers models dominate the field, achieving remarkable performance compared to older methods like hand-crafted rules and decision trees. However, these models are often less interpretable and they are unable to provide justification for their decisions. On the other hand, hand-crafted rules are inherently more interpretable but they achieve significantly lower performance. While there have been efforts to gain insights into the inner workings of DL models, it is still not entirely clear which are the factors that drive the model’s
behavior. In addition, both the fields of sentence compression and sentence splitting have received limited attention regarding interpretability. This highlights the need for further research that can bridge this gap by developing compression and splitting systems that offer high performance as well as transparency, trustworthiness and interpretability. 

\subsection{Complexity}
The majority of recent sentence compression and sentence splitting approaches are based on Transformer architectures, extensively trained on large annotated datasets. Even though these approaches demonstrate impressive results, they require heavy computational resources and also come with environmental consequences. The large size of Transformer models demands extensive computing power and resources, which can limit their use on resource-constrained applications or low-power devices. This issue highlights the importance of exploring more sustainable and lightweight alternatives while maintaining the same performance. The development of efficient architectures is an ongoing research that can significantly enhance the scalability of current systems. The balance between performance and efficiency remains an open research challenge in the entire field of NLP.

\subsection{Large Language Models in Sentence Compression and Splitting}
Despite the impressive performance of LLMs in various NLP tasks, their application in sentence compression and splitting has been limited. LLMs have the potential to significantly improve these tasks by providing deeper contextual understanding and enhancing adaptability to new domains and languages. There is strong reason to expect that LLMs could excel in sentence compression and splitting, particularly given their demonstrated success in abstractive text summarization~\cite{liu2023revisiting, zhang2024benchmarking}. Since sentence compression can be viewed as an extremely fine-grained form of summarization at the sentence level, and sentence splitting as a controlled form of content expansion and reorganization, the mechanisms used in summarization are directly applicable to both tasks.

Given this alignment, LLMs could substantially improve sentence compression and splitting by enabling more accurate decisions about which sentence elements are essential and how clauses should be reorganized. Furthermore, because LLMs generalize well across domains, they can support compression tasks in low-resource or specialized areas by generating high-quality synthetic training data or by acting as teacher models in distillation pipelines~\cite{Raffel2020}. This would allow lightweight models to retain the benefits of LLM-level understanding while remaining feasible for real-time or resource-constrained deployment.

However, given the computational demands and environmental impacts of these models, applying LLMs in this area requires a balance between their capabilities and resource limitations. For example, LLMs could be used strategically to generate training samples for low-resource domains or languages, while lighter models could be used for fine-tuning in these tasks. Additionally, multimodal sentence compression, which integrates information from both text and images, may particularly benefit from the contextual capabilities of LLMs, though achieving this efficiently remains an ongoing research challenge.

\section{Conclusions}
\label{sec:conclusions}
This paper provided a systematic review of sentence compression and splitting methods for resolving long sentences. We covered a wide spectrum of techniques, starting from traditional approaches, such as rule-based and statistical methods, and progressing towards more advanced solutions, such as RNNs, Transformers models and LLMs. We systematically categorized these methods into a comprehensive taxonomy and we provided a comparative analysis between existing models on widely used datasets for both tasks. In addition, we discussed the challenges and limitations associated with evaluating the effectiveness of these techniques. Supervised techniques, especially extractive ones, are the most commonly employed for sentence compression. With the advent of deep learning, there has been a significant increase in the development of abstractive sentence compression methods. Similarly, for sentence splitting, the landscape is dominated by supervised methods, with Transformers and RNN-based methods being the most effective ones. However, in simple scenarios, rule-based methods have also demonstrated their ability to recognize straightforward patterns. We observed a significant gap in weakly supervised and self-supervised techniques for sentence compression and unsupervised methods for sentence splitting. We also noted that, despite their potential, LLMs have not yet been extensively studied for these tasks. These unexplored areas provide the opportunity for further research on methods that do not rely on large annotated datasets, offering potential solutions to domains with limited data.

This survey highlights the importance of sentence compression and sentence splitting in improving the readability and comprehension of textual content, which is crucial in various applications and domains. Despite the significant progress that has been made in recent years, there remain several open issues for future research, such as addressing issues related to meaning preservation, optimizing the trade-off between sentence length reduction and information retention, and developing efficient and lightweight methods for resource-constrained devices. Further research on this topic offers a promising path to enhance current state-of-the-art methods and expand their usability across different applications.

\bibliographystyle{unsrt}  
\bibliography{bibl/bibliography, bibl/references, bibl/sample-base}

\end{document}